\pdfoutput=1

\documentclass[11pt]{article}

\usepackage[T1]{fontenc}

\usepackage[utf8]{inputenc}

\newcommand\camready[1]{}

\usepackage[table,xcdraw]{xcolor}
\usepackage[hyperref]{acl}
\usepackage{times}
\usepackage{amsmath}
\usepackage{latexsym}
\usepackage[capitalize,noabbrev]{cleveref}

\usepackage{graphicx}
\usepackage{multirow}
\usepackage{booktabs} 
\usepackage{subcaption}
\usepackage{siunitx}
\usepackage{xurl}  
\usepackage{enumitem}  
\usepackage{makecell}  
\usepackage{arydshln}
\usepackage{hyperref}
\usepackage{cleveref}
\crefformat{footnote}{#2\footnotemark[#1]#3}

\usepackage{microtype}

\usepackage{scalerel,stackengine}
\stackMath
\newcommand\reallywidehat[1]{%
\savestack{\tmpbox}{\stretchto{%
  \scaleto{%
    \scalerel*[\widthof{\ensuremath{#1}}]{\kern-.6pt\bigwedge\kern-.6pt}%
    {\rule[-\textheight/2]{1ex}{\textheight}}
  }{\textheight}%
}{0.5ex}}%
\stackon[1pt]{#1}{\tmpbox}%
}
\makeatletter
\def\adl@drawiv#1#2#3{%
        \hskip.5\tabcolsep
        \xleaders#3{#2.5\@tempdimb #1{1}#2.5\@tempdimb}%
                #2\z@ plus1fil minus1fil\relax
        \hskip.5\tabcolsep}
\newcommand{\cdashlinelr}[1]{%
  \noalign{\vskip\aboverulesep
          \global\let\@dashdrawstore\adl@draw
          \global\let\adl@draw\adl@drawiv}
  \cdashline{#1}
  \noalign{\global\let\adl@draw\@dashdrawstore
          \vskip\belowrulesep}}
\makeatother

\newcommand\scrolls{\textsc{Scrolls}}
	
\title{\scrolls{}: Standardized CompaRison Over Long Language Sequences}

\author{
Uri Shaham$^{\tau}$ \hspace{0.15cm} Elad Segal$^{\tau}$ \hspace{0.15cm} Maor Ivgi$^{\tau}$ \hspace{0.15cm} Avia Efrat$^{\tau}$ \\
\textbf{Ori Yoran}$^{\tau}$ \hspace{0.15cm} \textbf{Adi Haviv}$^{\tau}$ \hspace{0.15cm} \textbf{Ankit Gupta}$^{\beta}$ \hspace{0.15cm} \textbf{Wenhan Xiong}$^{\mu}$ \hspace{0.15cm} \textbf{Mor Geva}$^{\tau\alpha}$ \\
\textbf{Jonathan Berant}$^{\tau}$ \hspace{0.15cm} \textbf{Omer Levy}$^{\tau\mu}$ \\
\\
$^{\tau}$ The Blavatnik School of Computer Science, Tel Aviv University \\
$^{\alpha}$ Allen Institute for AI \\
$^{\beta}$ IBM Research \\
$^{\mu}$ Meta AI \\
}

\date{}

\begin{document}
\maketitle

\begin{abstract}

    NLP benchmarks have largely focused on \textit{short} texts, such as sentences and paragraphs, even though \textit{long} texts comprise a considerable amount of natural language in the wild.
    We introduce \scrolls{}, a suite of tasks that require reasoning over long texts.
    We examine existing long-text datasets, and handpick ones where the text is naturally long, while prioritizing tasks that involve synthesizing information across the input.
    \scrolls{} contains summarization, question answering, and natural language inference tasks, covering multiple domains, including literature, science, business, and entertainment.
    Initial baselines, including Longformer Encoder-Decoder, indicate that there is ample room for improvement on \scrolls{}.
        We make all datasets available in a unified text-to-text format and host a live leaderboard to facilitate research on model architecture and pretraining methods.\footnote{\label{website}\url{https://www.scrolls-benchmark.com}}

\end{abstract}

\section{Introduction}
\label{sec:introduction}

Standard benchmarks à la GLUE \cite{wang-etal-2018-glue, NEURIPS2019_4496bf24}, WMT \cite{barrault-etal-2019-findings, barrault-etal-2020-findings}, and SQuAD \cite{rajpurkar-etal-2016-squad, rajpurkar-etal-2018-know}, have driven progress in natural language processing of \textit{short} utterances.
However, a large portion of natural language is produced in the context of \textit{longer} discourses, such as books, articles, meeting transcripts, etc. 
To tackle the computational challenges associated with processing such long sequences, a plethora of new model architectures have recently emerged \cite{tay2020efficient, fournier2021practical},
\textit{without} establishing a standard scheme for evaluating them on long natural language problems.
Some long-context models are evaluated via language modeling perplexity, but this metric mostly captures model sensitivity to local, short-range patterns \cite{khandelwal-etal-2018-sharp, sun-etal-2021-long}.
Other studies rely on Long Range Arena \cite{tay2021long}, which is limited from a natural-language perspective, since only two of its datasets involve natural language, and those are artificially-elongated through byte tokenization.
To enable the research community to go beyond sentences and paragraphs,
we present a new benchmark, \textbf{\scrolls{}}: \textbf{S}tandardized \textbf{C}ompa\textbf{R}ison \textbf{O}ver \textbf{L}ong \textbf{L}anguage \textbf{S}equences. 

\begin{figure}[t]
    \centering
    \includegraphics[width=\columnwidth]{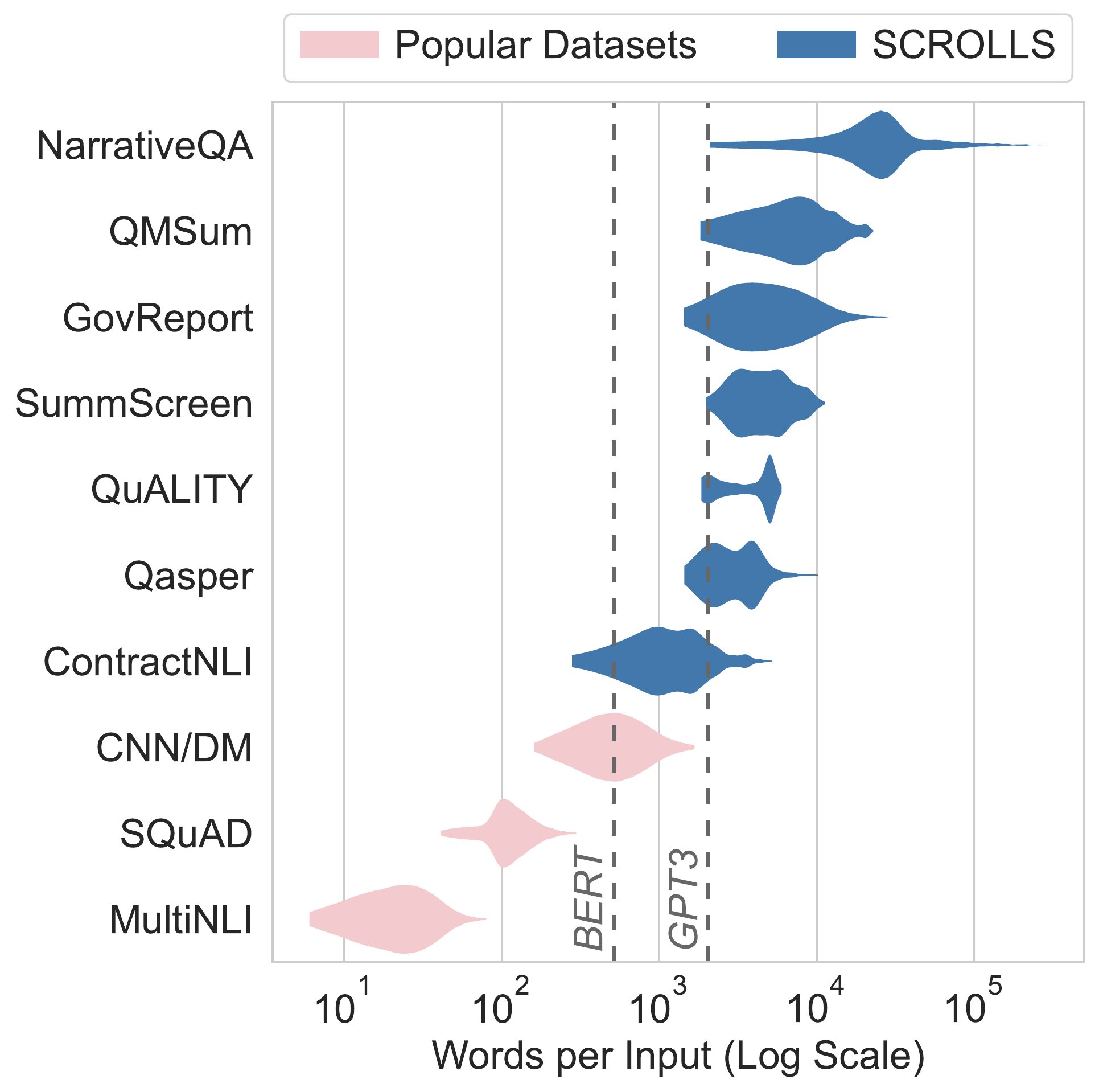}
    \caption{The distribution of words per input in \scrolls{} datasets (blue), alongside frequently-used NLP datasets (pink). Dashed vertical lines indicate the maximal sequence length (in tokens) of BERT \cite{devlin-etal-2019-bert} and GPT3 \cite{NEURIPS2020_1457c0d6}.} 
    \label{fig:input_length_dist}
\end{figure}

\scrolls{} incorporates multiple tasks (summarization, question answering, and natural language inference) over various domains (literature, meeting transcripts, TV shows, scientific articles, and more),
\textit{where each example's input typically contains thousands of words}.
We review the existing literature on long-text tasks and manually curate a subset of 7 datasets, prioritizing those that require contextualizing and abstracting information across multiple parts of the text.
We then clean and convert the data to a unified text-to-text format to enable the evaluation of a single model over all datasets.
Figure~\ref{fig:input_length_dist} shows that the texts in \scrolls{} datasets are substantially longer than commonly-used NLP benchmarks.
Moreover, our analysis reveals that, in \scrolls{}, critical information is spread out across longer distances within the input documents.

\scrolls{} is available via the Datasets library \cite{lhoest-etal-2021-datasets} or direct download on its website, which hosts a live leaderboard that accepts submissions and automatically evaluates them against private test sets.
By producing a single aggregate score, in addition to individual dataset scores, \scrolls{} can serve as an evaluation platform for future approaches to processing long text, whether by new pretraining schemes, novel transformer architectures and alternatives, or even retrieval-based methods.
We provide initial baselines for \scrolls{} using two transformer models, BART \cite{lewis-etal-2020-bart}, and its length-efficient variant, Longformer Encoder-Decoder \cite{beltagy2020longformer}.
Our experiments indicate that \scrolls{} poses a formidable challenge for these models, leaving much room for the research community to improve upon.

\section{Background: Contemporary Evaluation of Long-Text Models}
\label{sec:RelatedWork}

While transformers \cite{NIPS2017_3f5ee243} are the current go-to architecture for building state-of-the-art models in NLP, they present a computational challenge when it comes to long sequences due to the $O(n^2)$ complexity of self-attention, where $n$ is the sequence's length. 
To address this problem, a wide variety of efficient alternatives and approximations have been proposed over the past couple of years \cite{tay2020efficient, fournier2021practical}.
Much of these novel architectures were developed concurrently, leading to somewhat of a ``Wild West'' when it comes to model evaluation, making cross-model comparison challenging.
Roughly speaking, we can cluster the more prominent evaluation methodologies into three categories: language modeling, Long-Range Arena, and summarization.

The language modeling community typically uses perplexity to measure how well models predict the next token,
a practice that has been adopted by several works on efficient transformer architectures \cite{roy-etal-2021-efficient, choromanski2020rethinking, tay2020synthesizer, peng2021random}.
However, using perplexity to evaluate a model's \textit{long-range} abilities is currently under scrutiny.
A growing amount of literature shows that predicting the next token is mostly a local task that does not require modeling long-range dependencies \cite{khandelwal-etal-2018-sharp, sun-etal-2021-long}, and that masking or down-weighting distant tokens can actually \textit{improve} perplexity \cite{press-etal-2021-shortformer, press2021train}.

A more recent approach to standardizing long-sequence model evaluation is the Long Range Arena (LRA) \cite{tay2021long}.
It incorporates 5 classification datasets: 
byte-level sentiment analysis (IMDB) and document relatedness (ACL Anthology);
path-finding (Pathfinder) and image classification (CIFAR-10) over 1-dimensional pixel sequences;
and executing a list of mathematical operations (ListOps).
Of those, two involve visual reasoning, and one is a synthetic mathematical language (ListOps), leaving only two natural language datasets (sentiment analysis and document relatedness).
The multi-modal nature of LRA makes it inappropriate as a testbed for pretrained language models, limiting its relevance for NLP.
Moreover, LRA artificially inflates natural language sequences via byte tokenization, and truncates each example at 4,000 bytes, which is equivalent to less than 1,000 words.
This exempts models from coping with the complex long-range dependencies that exist in naturally long texts.

The third practice uses summarization tasks to evaluate long-sequence models.
The most popular datasets use abstracts of academic papers on arXiv and PubMed \cite{cohan-etal-2018-discourse} as summaries.
Other summarization datasets, however, are less frequently used, biasing the evaluation towards academic domains.
\scrolls{} includes summarization as one of its main tasks, selecting datasets from several different domains to increase diversity.

\section{The \scrolls{} Benchmark}

\begin{table*}[th]
\small
\centering
\begin{tabular}{@{}llllrrr@{}}
\toprule
\multirow{2}{*}{\textbf{Dataset}} & \multirow{2}{*}{\textbf{Task}} & \multirow{2}{*}{\textbf{Domain}} & \multirow{2}{*}{\textbf{Metric}} & \multicolumn{2}{c}{\textbf{Avg \#Words}} &  \multirow{2}{*}{\textbf{\#Examples}}  \\
& & &  & \textbf{Input} & \textbf{Output} &   \\
\midrule
GovReport \cite{huang-etal-2021-efficient} & Summ & Government & ROUGE & 7,886 & 492.5 & 19,402 \\
SummScreenFD \cite{chen2021summscreen} & Summ & TV & ROUGE & 5,598 & 99.6 & 4,348 \\
QMSum \cite{zhong-etal-2021-qmsum} & QB-Summ & Meetings & ROUGE & 9,497 & 69.7 & 1,810 \\
Qasper \cite{dasigi-etal-2021-dataset} & QA & Science & F1 & 3,629 & 11.4 & 5,692 \\
NarrativeQA \cite{kocisky-etal-2018-narrativeqa} & QA & Literature, Film & F1 & 51,653 & 4.6 & 71,187 \\
QuALITY \cite{pang2021quality} & MC-QA & Literature, Misc & EM & 4,193 & 10.3 & 6,737 \\
ContractNLI \cite{koreeda2021contractnli} & NLI & Legal & EM & 1,706 & 1.4 & 10,319 \\
\bottomrule
\end{tabular}
\caption{An overview of the datasets in \scrolls{} and their statistics. \textit{Summ} refers to summarization, \textit{QB-Summ} means query-based summarization, and \textit{MC-QA} abbreviates multiple-choice question answering. The number of examples includes train, validation, and test sets.}
\label{tab:datasets_statistics}
\end{table*}

\scrolls{} aims to challenge a model's ability to process long texts in the wild, and therefore focuses on discourses that are \textit{naturally} long, encompassing domains such as literature, TV show scripts, scientific articles, and more.
We review the datasets in existing literature, seeking ones that challenge models not only by the length of each input, but also by the need to process long-range dependencies across different sections.
At the same time, we strive to maintain a diversity of tasks, covering summarization and query-based summarization, open ended and multiple-choice question answering, as well as natural language inference.

Through this curation process, we handpick 7 datasets, and process them into a uniform text-to-text format.
Table~\ref{tab:datasets_statistics} provides an overview of the datasets included in \scrolls{}.
Figure~\ref{fig:ssfd_example} and Figure~\ref{fig:quality_example} show two examples from \scrolls{} datasets SummScreenFD and QuALITY, demonstrating how contextualizing and synthesizing information over long ranges of text is paramount to addressing the challenges in the benchmark.

\subsection{Datasets}

We survey the 7 datasets in \scrolls{}, and elaborate how the original data was collected.

\paragraph{GovReport} \cite{huang-etal-2021-efficient}:
A summarization dataset of reports addressing various national policy issues published by the Congressional Research Service\footnote{https://crsreports.congress.gov/} and the U.S. Government Accountability Office,\footnote{https://www.gao.gov/} where each document is paired with an expert-written executive summary.
The reports and their summaries are longer than their equivalents in other popular long-document summarization datasets; for example, GovReport's documents are approximately 1.5 and 2.5 times longer than the documents in arXiv and PubMed \cite{cohan-etal-2018-discourse}, respectively.

\paragraph{SummScreenFD} \cite{chen2021summscreen}:
A summarization dataset in the domain of TV shows (e.g. Friends, Game of Thrones).
Given a transcript of a specific episode, the goal is to produce the episode's recap.
The original dataset is divided into two complementary subsets, based on the source of its community contributed transcripts. 
For \scrolls{}, we use the ForeverDreaming (FD) subset,\footnote{http://transcripts.foreverdreaming.org} as it incorporates 88 different shows, 
making it a more diverse alternative to the TV MegaSite (TMS) subset,\footnote{http://tvmegasite.net/} which has only 10 shows.
Community-authored recaps for the ForeverDreaming transcripts were collected from English Wikipedia and TVMaze.\footnote{https://www.tvmaze.com/}

\paragraph{QMSum} \cite{zhong-etal-2021-qmsum}:
A query-based summarization dataset, consisting of 232 meetings transcripts from multiple domains and their corresponding summaries. 
The corpus covers academic group meetings at the International Computer Science Institute \cite{1198793},\footnote{https://groups.inf.ed.ac.uk/ami/icsi/index.shtml} industrial product meetings for designing a remote control \cite{10.1007/11677482_3}, and committee meetings of the Welsh\footnote{https://record.assembly.wales} and Canadian\footnote{https://www.ourcommons.ca/Committees/en/Home} Parliaments, dealing with a variety of public policy issues.
Annotators were tasked with writing queries about the broad contents of the meetings, as well as specific questions about certain topics or decisions, while ensuring that the relevant text for answering each query spans at least 200 words or 10 turns.


\paragraph{Qasper} \cite{dasigi-etal-2021-dataset}:
A question answering dataset over NLP papers filtered from the Semantic Scholar Open Research Corpus (S2ORC) \cite{lo-etal-2020-s2orc}.
Questions were written by NLP practitioners after reading only the title and abstract of the papers, while another set of NLP practitioners annotated the answers given the entire document.
Qasper contains abstractive, extractive, and yes/no questions, as well as unanswerable ones. 

\begin{figure}[t]
\small
\centering
{\setlength{\extrarowheight}{3pt}%
\begin{tabular}{|p{0.94\columnwidth}|} 
\hline
\textcolor[HTML]{AFAFAF}{Penny returns from visiting family in Nebraska, but mentions while picking up mail from Leonard that most of her relatives became sick. Sheldon, a germophobe according to Leonard, freaks out and becomes sick, becoming demanding on top of his already obnoxious personality. \textcolor[HTML]{000000}{Familiar with Sheldon being sick, Leonard and the guys hide from him at a Planet of the Apes series marathon, leaving Penny to care for Sheldon.} However, Leonard breaks his glasses in the cinema and has to retrieve his spare pair from the apartment, piloted by Howard and Raj using a laptop, an endoscope, and a Bluetooth helmet camera worn by the short-sighted Leonard. Penny intercepts him and abandons him to his fate with Sheldon. Leonard tries to escape, but runs into a wall and nearly knocks himself out. In the end, injured Leonard and sick Sheldon sit miserably on the couch.}
\vspace{2pt}
\\
\multicolumn{1}{|c|}{\textcolor[HTML]{AFAFAF}{------ \texttt{{Transcript}} ------}} \\

\textcolor[HTML]{3078BE}{...[1,032 words]...}

\textcolor[HTML]{AFAFAF}{Howard: Hello.}

\textcolor[HTML]{000000}{Sheldon: Howard, I'm sick.}

\textcolor[HTML]{3078BE}{...[40 words]...}

\textcolor[HTML]{AFAFAF}{Howard: It's my own fault, I forgot the protocol we put in place after the great ear infection of '06.}

\textcolor[HTML]{000000}{Leonard: You call Koothrappali, we need to find a place to lay low for the next eighteen to twenty four hours.}

\textcolor[HTML]{AFAFAF}{Howard: Stand by. Ma, can my friends come over?}

\textcolor[HTML]{AFAFAF}{Howard's Mother: I just had the carpets steamed.}

\textcolor[HTML]{000000}{Howard: That's a negatory. But there's a Planet of the Apes marathon at the New Art today.
}

\textcolor[HTML]{000000}{Leonard: Five movies, two hours apiece. It's a start.}

\textcolor[HTML]{3078BE}{...[660 words]...}

\textcolor[HTML]{AFAFAF}{Sheldon: Based on what happened next, I assume it means "would you like an enema?"}

\textcolor[HTML]{000000}{Penny: Okay, sweetie, I'll take care of you, what do you need?}

\textcolor[HTML]{3078BE}{...[766 words]...}

\textcolor[HTML]{000000}{Penny: You deliberately stuck me with Sheldon.}

\textcolor[HTML]{AFAFAF}{Leonard: Well, I had to, you see what he's like.}

\textcolor[HTML]{3078BE}{...[142 words]...}
\vspace{2pt}
\\
\hline
\end{tabular}}
\caption{An example from the SummScreenFD summarization dataset, where the task is to generate the recap (top paragraph) given the episode's script. In this example, the information required to compose the third sentence in the recap (highlighted) is scattered across several snippets throughout the transcript.}
\label{fig:ssfd_example}
\end{figure}

\begin{figure}[t]
\small
\centering
{\setlength{\extrarowheight}{3pt}%
\begin{tabular}{|p{0.94\columnwidth}|} 
\hline
The text says ``The expert frowned horribly.'' What makes the expert's smile so horrible?

\textcolor[HTML]{AFAFAF}{(A) The frown indicates that he's close to detecting Korvin's true motivations.}

\textcolor[HTML]{AFAFAF}{(B) The frown indicates that he knows that Korvin switched the wires on the lie detector.}

\textcolor[HTML]{AFAFAF}{(C) The frown is a signal to the Ruler that Korvin is lying.}

(D) The frown is physically horrible because the Tr'en have fifty-eight, pointed teeth.
\vspace{2pt}
\\
\multicolumn{1}{|c|}{\textcolor[HTML]{AFAFAF}{------ \texttt{Story} ------}} \\
\textcolor[HTML]{3078BE}{...[607 words]...}

\textcolor[HTML]{AFAFAF}{It was a ritual, Korvin had learned. ``You are of the Tr'en,'' he replied. \textcolor[HTML]{000000}{The green being nodded. ``I am Didyak of the Tr'en,'' he said.}}

\textcolor[HTML]{3078BE}{...[257 words]...}

\textcolor[HTML]{000000}{Didyak beamed at him. The sight was remarkably unpleasant, involving as it did the disclosure of the Tr'en fifty-eight teeth, mostly pointed.} \textcolor[HTML]{AFAFAF}{Korvin stared back impassively. ``I have been ordered to come to you,'' Didyak said, ``by the Ruler. The Ruler wishes to talk with you.''}

\textcolor[HTML]{3078BE}{...[1,366 words]...}

\textcolor[HTML]{AFAFAF}{``They can be treated mathematically,''} \textcolor[HTML]{000000}{one of the experts, a small emerald-green being,} \textcolor[HTML]{AFAFAF}{told Korvin thinly. ``Of course, you would not understand the mathematics.''}

\textcolor[HTML]{3078BE}{...[33 words]...}

\textcolor[HTML]{000000}{The expert frowned horribly, showing all of his teeth.} \textcolor[HTML]{AFAFAF}{Korvin did his best not to react. ``Your plan is a failure,'' the expert said, ``and you call this a good thing.''}

\textcolor[HTML]{3078BE}{...[1,808 words]...}
\vspace{2pt}
\\
\hline
\end{tabular}}
\caption{An example from the QuALITY dataset, where the task is to answer multiple-choice questions about a given story or document. In this example, answering the question correctly requires reasoning over four different snippets that are separated by long token sequences.}

\label{fig:quality_example}
\end{figure}

\paragraph{NarrativeQA} \cite{kocisky-etal-2018-narrativeqa}:
An established question answering dataset over entire books from Project Gutenberg\footnote{\label{gutenberg}http://www.gutenberg.org} and movie scripts from different websites.\footnote{http://www.imsdb.com, http://www.dailyscript.com/, http://www.awesomefilm.com}
Annotators were given summaries of the books and scripts obtained from Wikipedia, and asked to generate question-answer pairs, resulting in about 30 questions and answers for each of the 1,567 books and scripts.
They were encouraged to use their own words rather then copying, and avoid asking yes/no questions or ones about the cast.
Each question was then answered by an additional annotator, providing each question with two reference answers (that may be identical).

\paragraph{QuALITY} \cite{pang2021quality}:
A multiple-choice question answering dataset over stories and articles sourced from Project Gutenberg,\cref{gutenberg} the Open American National Corpus \cite{Fillmore1998AnAN, ide-suderman-2004-american}, and more.
Experienced writers wrote questions and distractors, and were incentivized to write answerable, unambiguous questions such that in order to correctly answer them, human annotators must read large portions of the given document. 
To measure the difficulty of their questions, Pang et al. conducted a speed validation process, where another set of annotators were asked to answer questions given only a short period of time to skim through the document.
As a result, 50\% of the questions in QuALITY are labeled as \textit{hard}, i.e. the majority of the annotators in the speed validation setting chose the wrong answer.

\paragraph{Contract NLI} \cite{koreeda2021contractnli}:
A natural language inference dataset in the legal domain.
Given a non-disclosure agreement (NDA, the premise), the task is to predict whether a particular legal statement (the hypothesis) is entailed, not entailed (neutral), or cannot be entailed (contradiction) from the contract.
The NDAs were manually picked after simple filtering from the Electronic Data Gathering, Analysis, and Retrieval system (EDGAR)\footnote{https://www.sec.gov/Archives/edgar/Oldloads} and Google.
The dataset contains a total of 607 contracts and 17 unique hypotheses, which were combined to produce the dataset's 10,319 examples.

\subsection{Preprocessing}

\paragraph{Data Cleansing}
As part of the curation process, we examine each dataset and clean or filter examples to ensure high quality data.
In GovReport, we discard all examples where the report's length (in words) is less than twice the summary, or more than 1,000 times the summary, as well as examples where the summary exists verbatim in the report. This process removes 64 examples from the original dataset.
In Qasper, we discard all papers that have less than 8,192 characters, removing a total of 176 questions over 63 papers, which appear to be of lower quality.
In NarrativeQA, we locate markers indicating the start and end of the actual story, and use them to remove excess metadata such as licenses, HTML headers, etc.

\paragraph{Unified Format}
We reformulate every dataset in \scrolls{} as a sequence-to-sequence task to allow for a simple unified input-output format.
When a query is given in addition to the raw text (as in QMSum, Qasper, NarrativeQA, QuALITY, and ContractNLI), we prepend it to the text, using two newlines as a natural separator.
For the multiple-choice dataset QuALITY, we also provide all four answer candidates as part of the query.
For the summarization datasets, GovReport and SummScreenFD, we use only the original documents as input.
Some datasets (Qasper and NarrativeQA) contain multiple target outputs for each input; we split them into separate instances for training and development. For test, we score each prediction with every valid answer independently, and then merge the scores of identical inputs by taking the maximum of those scores.
Table \ref{tab:data_examples} in Appendix~\ref{appx:dataset_format} provides an example from each \scrolls{} dataset. 



\subsection{Evaluation}

Each dataset is split into training, validation, and test sets based on the original dataset splits.
In \scrolls{}, test set outputs are kept private, and only the inputs are publicly available.
When evaluating a model, users must submit their model's outputs for \textit{all} test sets via the \scrolls{} website.
Once a model is submitted, we compute the average performance metric across all datasets to provide the submission with a single aggregate \scrolls{} score.
We employ three different evaluation metrics across \scrolls{} datasets: ROUGE for summarization tasks (GovReport, SummScreenFD, and QMSum), unigram overlap (F1) for question answering (Qasper and NarrativeQA), and exact match (EM) for multiple-choice (QuALITY) and classification (ContractNLI) tasks.
The official evaluation script is available online.\footnote{\label{github}\url{https://github.com/tau-nlp/scrolls}}

\paragraph{ROUGE}
We use three flavors of ROUGE \cite{lin-2004-rouge} to measure the overlap between the system-generated output and the reference: unigram overlap (ROUGE-1), bigram overlap (ROUGE-2), and the longest overlapping subsequence (ROUGE-L).
Both system output and reference are normalized by lowercasing and converting all non-alphanumeric characters to whitespaces, followed by whitespace tokenization.
We compute the geometric mean of the three scores (ROUGE-1/2/L) to produce a single score per dataset, which is used to calculate the final \scrolls{} score.\footnote{We discuss the limitations of using ROUGE to evaluate summarization in Section \ref{sec:limitations}.}

\paragraph{F1}
Similar to ROUGE-1, the F1 metric calculates unigram overlap.
The key difference is that both reference and system output strings are normalized slightly differently; in addition to lowercasing and punctuation removal, stopwords are also discarded, following the practice of SQuAD \cite{rajpurkar-etal-2016-squad} and other question-answering datasets \cite{fisch-etal-2019-mrqa}.
Both Qasper and NarrativeQA contain questions with more than one reference answer; for each such example, we take the maximal F1 score over all of its reference answers.

\paragraph{EM}
Exact match normalizes the output strings using the same procedure as F1 (lowercasing, removing punctuation and stopwords, and normalizing whitespaces), and then compares whether the two normalized strings are identical.
For QuALITY, we calculate EM over the entire test set (EM-T), and also EM over its subset of \textit{hard} questions (EM-H), as defined in the original dataset.
For computing the final \scrolls{} score, however, we only use the EM value calculated over the full test set (EM-T).

\section{Quantitative Analysis}

Length alone is not enough to make \scrolls{} a challenging benchmark.
Here, we provide a quantitative analysis that suggests that producing the correct output for a \scrolls{} task typically requires fusing different parts of the input that are often hundreds and even thousands of words apart.
This analysis complements the qualitative inspection of examples from \scrolls{}, as shown in Figure~\ref{fig:ssfd_example} and Figure~\ref{fig:quality_example}, and further discussed in Appendix~\ref{appx:qualitative_analysis}.

\paragraph{Methodology}
Each example in \scrolls{} consists of a textual input and output.\footnote{For the purposes of this analysis, we consider the output of ContractNLI to be the hypothesis, and the input to be the premise. For question answering datasets, the question is omitted.}
Given a specific input-output pair, we measure the example's \textit{spread} by computing the standard deviation between the locations of output bigrams in the input.\footnote{This metric is inspired by the analysis of \citet{huang-etal-2021-efficient} for GovReport, which also used bigram statistics.}
Specifically, we represent the output string as a set of bigrams, and locate the \textit{first occurrence} of each bigram in the input (if exists); we then compute the standard deviation between these locations (where a bigram is represented by the position of its first word in the input).
Now that we have an example-level measure of spread, we can plot an entire dataset's spread on a histogram, and compare different datasets.

\begin{figure*}[t]
\centering
\begin{subfigure}{.5\textwidth}
  \centering
  \includegraphics[width=\columnwidth]{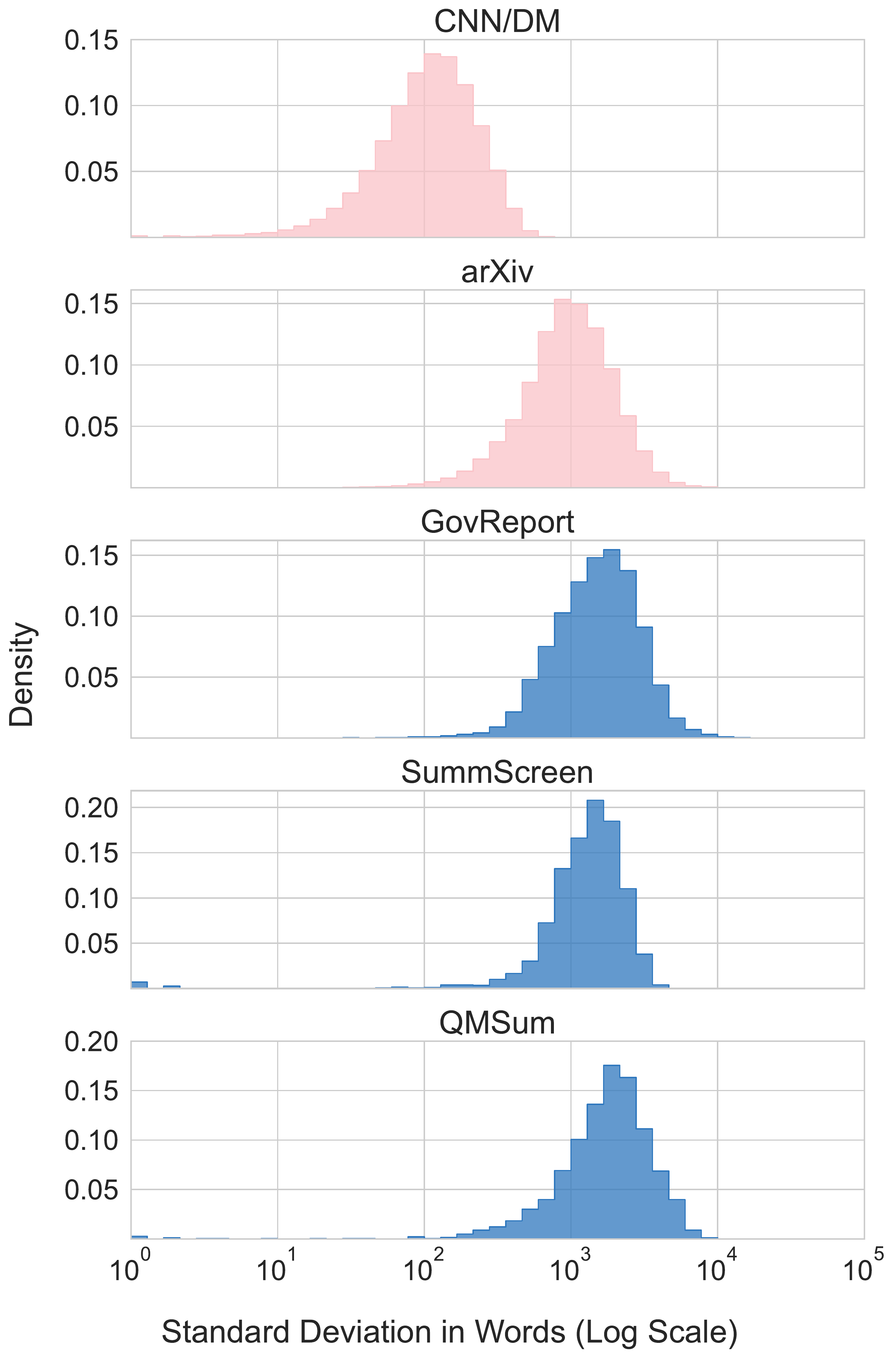}
  \caption{Summarization datasets}
  \label{fig:summ}
\end{subfigure}%
\begin{subfigure}{.5\textwidth}
  \centering
  \includegraphics[width=\columnwidth]{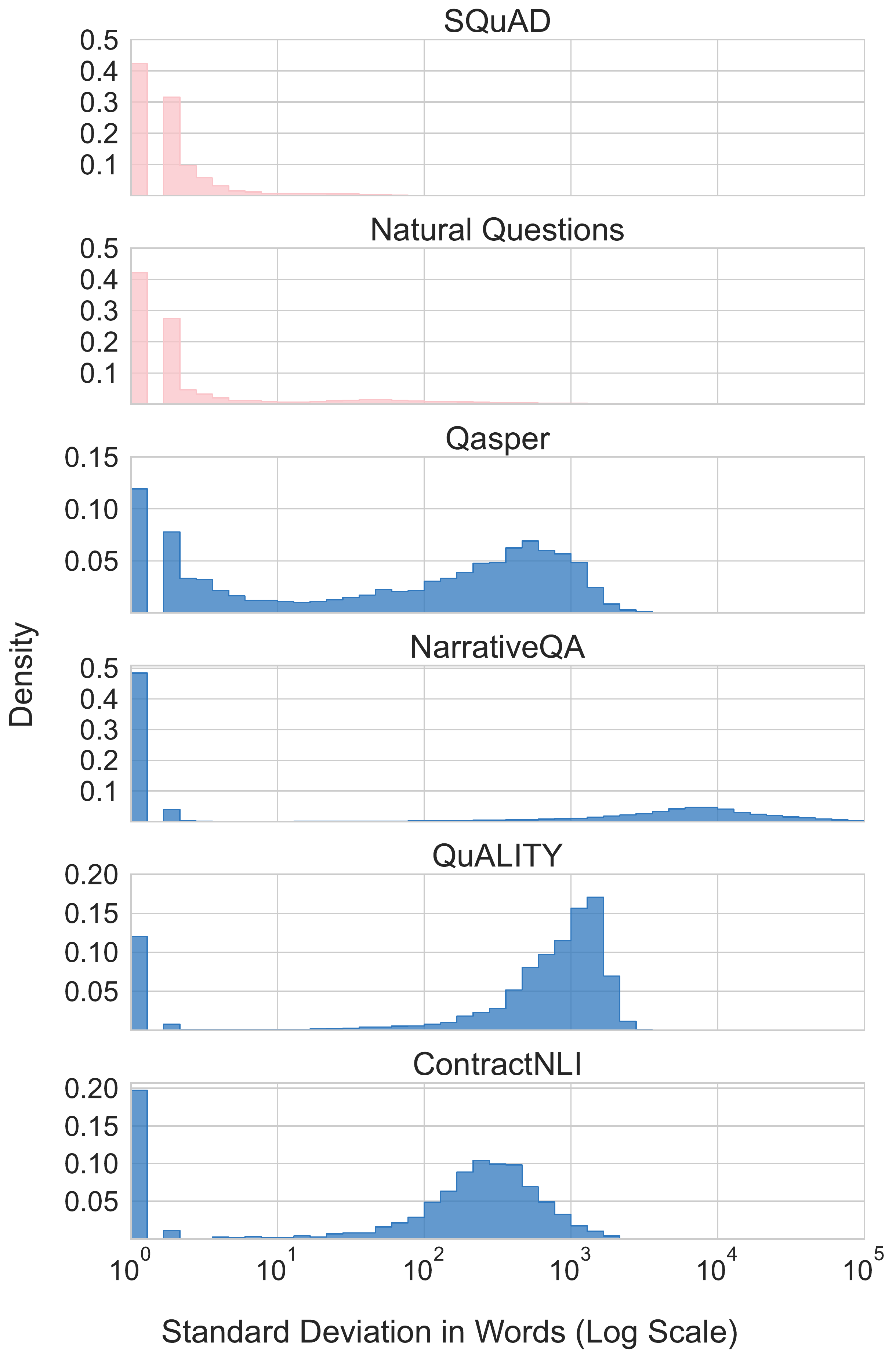}
  \caption{QA and NLI datasets}
  \label{fig:qa}
\end{subfigure}
\caption{The spread of reference-text bigrams in the input texts, measured by the standard deviation of the position of each bigram's first occurrence in the input document. \scrolls{} datasets (blue), other popular datasets (pink).}
\label{fig:bigrams}
\end{figure*}

\paragraph{Summarization Datasets}
Figure \ref{fig:summ} compares the three summarization datasets in \scrolls{} to the canonical CNN/DM summarization dataset \cite{NIPS2015_afdec700}, as well as arXiv \cite{cohan-etal-2018-discourse}, which has been used to evaluate long-sequence models.
We observe that the reference bigrams are spread out across much larger distances in \scrolls{} than in CNN/DM, and by a factor of 1.5 to 2 times more than arXiv on average.

\paragraph{QA \& NLI Datasets}
Figure \ref{fig:qa} compares the remaining four datasets in \scrolls{}, which typically have shorter outputs, to the popular SQuAD \cite{rajpurkar-etal-2016-squad} and Natural Questions \cite{kwiatkowski-etal-2019-natural} datasets.\footnote{We use the version of Natural Questions that takes entire Wikipedia articles as inputs and short answers as outputs.}
While the answer bigrams in SQuAD and Natural Questions typically spread across distances of under 5 words, the output bigrams in \scrolls{} datasets are usually separated by hundreds of words.
NarrativeQA also seems to contain many examples where the answer bigrams cluster close together, but also a significant subset of examples where the answer's bigrams are dispersed across huge distances.

\section{Experiments}

We conduct experiments to evaluate the ability of mainstream models to handle the various long text challenges presented by \scrolls{}.
Our code is based on the Transformers library \cite{wolf-etal-2020-transformers}, and is available online.\cref{github}

\subsection{Baselines}

We finetune two pretrained transformer variants as baselines, as well as naive heuristic baselines to establish the floor performance on each task.
Hyperparameters are detailed in Appendix~\ref{appx:hyperparameters}.

\paragraph{BART} 
As a standard transformer baseline, we use the pretrained BART-base\footnote{https://huggingface.co/facebook/bart-base} model \cite{lewis-etal-2020-bart}. 
BART is a transformer encoder-decoder pretrained by reconstructing noised texts, which achieved state-of-the-art results on several summarization datasets when released.
BART was pretrained on sequences of up to 1,024 tokens; we therefore truncate all inputs by retaining only their 1,024-token prefix.
To examine the effect of available input length, we also consider truncating BART's inputs at 256 and 512 tokens.

\paragraph{Longformer Encoder-Decoder (LED)}
We experiment with LED-base,\footnote{https://huggingface.co/allenai/led-base-16384} the encoder-decoder version of the efficient transformer architecture Longformer \cite{beltagy2020longformer}.
Longformer avoids computing quadratic-complexity attention via sliding-window attention, where each word only attends to a constant number of nearby tokens, in addition to a few tokens that compute global attention over the entire input.
LED is initialized with BART's parameters, without further pretraining.
In our experiments, we use a sliding window of 1,024 tokens, and restrict the total input length to 16,384 tokens via truncation, following Beltagy et al.
We also experiment with maximum sequence lengths of 1,024 and 4,096 tokens.
While the original work on LED selects the globally-attending tokens on a per-task basis, we follow their summarization setting throughout all tasks, which enables global attention only for the first token.

\paragraph{Heuristic Baselines}
We use simple heuristics to find the lower bound of performance on each dataset.
For most datasets, we use the fixed-length prefix heuristic, akin to the LEAD baseline in the summarization literature.
Specifically, we compute the average output-input length ratio $\rho$ over the training set (in characters), and then produce the first $\rho \cdot n$ characters from the given input at inference time (where $n$ is the input's length).
For QuALITY, we use the majority class (which is just above one quarter).
For ContractNLI, we use the per-hypothesis majority class, as the same 17 hypotheses are shared across all documents.

\begin{table*}[th]
\small
\centering
\begin{tabular}{@{}l@{}rcccccccc@{}}
\toprule
\multirow{2}{*}{\textbf{Model}} & \multirow{2}{*}{\textbf{~(Input)}} & \textbf{GovRep} & \textbf{SumScr} &  \textbf{QMSum} & \textbf{Qspr}  & \textbf{Nrtv}  & \textbf{QALT}  & \textbf{CNLI} & \multirow{2}{*}{\textbf{Avg}}  \\
&  & ROUGE-1/2/L  &  ROUGE-1/2/L  &  ROUGE-1/2/L  & F1  & F1 & EM-T/H &  EM  & \\
\midrule
Naive & ~~~~~~~~- & 45.3 / 17.9 / 20.8 & 19.6 / 1.8 / 11.0 & 14.2 / 2.0 / 9.3 & 3.4 & 1.5 & 25.2 / 26.1 & 66.0 & \textit{19.35} \\ 
\midrule 
\multirow{3}{*}{BART} 
& ~~~~256 & 41.9 / 14.2 / 20.3 & 24.5 / 3.8 / 15.3 & 29.9 / 8.3 / 20.4 & 23.3 & 14.0 & 26.0 / 25.8 & 69.8 & \textit{26.35} \\ 
& ~~~~512 & 45.6 / 16.9 / 21.8 & 26.3 / \textbf{5.1} / 16.2 & 29.5 / 8.2 / 20.1 & 24.7 & 14.5 & \textbf{26.8} / \textbf{27.4} & 71.6 & \textit{27.58} \\ 
& ~~1024 & 47.9 / 18.6 / 22.7 & \textbf{27.2} / 4.9 / \textbf{16.7} & \textbf{30.2} / \textbf{8.7} / \textbf{20.7} & 26.3 & 15.4 & 26.0 / 25.9 & \textbf{77.4} & \textit{29.01} \\ 
\midrule 
\multirow{3}{*}{LED} 
& ~~1024 & 40.9 / 16.1 / 23.1 & 22.7 / 3.6 / 15.1 & 24.6 / 6.5 / 19.0 & 24.4 & 15.2 & 26.6 / 27.2 & 73.4 & \textit{27.06} \\ 
& ~~4096 & 52.5 / 23.3 / 26.8 & 23.0 / 4.1 / 15.1 & 26.6 / 6.9 / 19.9 & 25.0 & 16.3 & 26.6 / 27.3 & 71.5 & \textit{28.30} \\ 
& 16384 & \textbf{56.2} / \textbf{26.6} / \textbf{28.8} & 24.2 / 4.5 / 15.4 & 25.1 / 6.7 / 18.8 & \textbf{26.6} & \textbf{18.5} & 25.8 / 25.4 & 71.5 & \textbf{\textit{29.16}} \\
\bottomrule
\end{tabular}
\caption{Baseline results on \scrolls{}, using naive heuristics, BART, and Longformer Encoder-Decoder (LED), and various input length limits. The final \scrolls{} score (Avg) is computed by averaging over each dataset's overall performance score. For QuALITY (QALT), we use the EM score calculated over the full test set (EM-T), without up-weighting the performance on the hard subset (EM-H). For datasets evaluated with ROUGE, we aggregate the different ROUGE scores via geometric mean to produce a single score per dataset, which is then used when calculating the final average \scrolls{} score.}
\label{tab:baseline_results}
\end{table*}

\subsection{Results}

Table \ref{tab:baseline_results} shows the baselines' performance on \scrolls{}.
A few trends are apparent:

\paragraph{More Context Improves Performance}
We experiment with three context lengths for each model.
As the model receives more context, its average \scrolls{} score increases.
For BART, increasing the input length from 256 tokens to 1,024 increases performance by 2.66 points, while LED grows by 2.1 points when enlarging its maximal sequence length from 1,024 tokens to 16,384.
This trend is relatively consistent across datasets for BART, but less so for LED (e.g., QMSum and ContractNLI).

\paragraph{BART versus LED}
Although LED does achieve the highest \scrolls{} score when given 16,384 tokens per sequence, BART arrives within 0.15 points of the top score \textit{despite being limited to only 1,024 tokens}.
This is surprising, given the substantial difference in input lengths.
Moreover, when controlling for the number of input tokens, BART outperforms LED by almost two points, suggesting that LED might be under-optimized.
Inspecting the dataset-level results reveals that LED (16k) significantly outperforms BART (1k) in two datasets, GovReport and NarrativeQA, which are coincidentally the largest datasets in \scrolls{} by number of examples.
Thus, it is possible that since LED is initialized with BART's parameters (without long-text pretraining), it requires a substantial amount of data and fine-tuning to adapt its parameters to sliding window attention and potentially longer inputs.

Overall, our experiments highlight the  importance of measuring not only whether an architecture can efficiently process long sequences, but also whether it can effectively model their semantics -- precisely what \scrolls{} is designed to do.

\paragraph{How Far is \scrolls{} from being Solved?}
The heuristic baselines set a lower bound average score of 19.35, which the model baselines are able to improve upon by 7 to 10 points.
While it is difficult to establish an accurate human performance ceiling on \scrolls{}, especially when considering the summarization datasets, we do have some indicators that it is probably much higher than the current baselines.
\citet{dasigi-etal-2021-dataset} study a subset of Qasper that has multiple annotated answers, and find their overlap to be 60.9\% F1, more than double our best baseline.
Likewise, human agreement on QuALITY was measured at 93.5\% EM \cite{pang2021quality}.
We also compute the inter-annotator agreement (F1) on NarrativeQA's test set (where each question has two answers), arriving at around 58.7\% F1, compared to our best baseline of 18.5\% F1.
Overall, it seems that contemporary off-the-shelf models struggle with these tasks, challenging future work to make progress on \scrolls{}.

\section{Conclusion}

We propose a new benchmark that places the spotlight on naturally long texts and their intricacies.
\scrolls{} fills a current gap around evaluating efficient transformer architectures and their alternatives on natural language tasks, and at the same time provides a testing ground for new pretraining schemes that target long language sequences.
We hope that \scrolls{} inspires the NLP community to go beyond single sentences and paragraphs, and meet the challenges of processing and reasoning over longer discourses.

\section{Limitations}
\label{sec:limitations}

The main limitation of \scrolls{} is the evaluation of long output texts, specifically in summarization.
Since ROUGE only accounts for $n$gram overlaps,
it might downvalue paraphrases of the reference summary that contain the same semantic content.
Establishing unbiased, automated metrics for long generations that correlate well with humans judgments is an emerging field of research,
and we may indeed decide to replace or complement ROUGE with model-based evaluation in the future.

A second limitation is that \scrolls{} is monolingual. Model evaluation over languages other than English has major significance, affecting the usage of language processing technology in applications worldwide. \scrolls{} is limited in that sense, but takes an initial step in standardizing evaluation over long text in general. A natural future direction is establishing benchmarks focusing on other languages as well.
 \section*{Acknowledgements}

This research was supported in part by the Yandex Initiative for Machine Learning, the Tel Aviv University Data Science Center, the Shashua Fellowship, the Alon Scholarship, the 2021 Scholarship in Computer Science granted by the Séphora Berrebi Foundation,
and the European Research Council (ERC) under the European Union Horizons 2020 research and innovation programme (grant ERC DELPHI 802800).

\bibliographystyle{acl_natbib}
\bibliography{anthology,custom}

\appendix
\newpage
\section{Dataset Format}
\label{appx:dataset_format}

Table \ref{tab:data_examples} shows an example from each dataset in \scrolls{}.

\section{Dataset Splits}
\label{appx:test_stats}

\begin{table}[h]
\small
\centering
\begin{tabular}{@{}lrrr@{}}
\toprule
\multirow{2}{*}{\textbf{Dataset}} & \multicolumn{3}{c}{\textbf{\#Examples}}   \\
& \textbf{Train} & \textbf{Valid} & \textbf{Test}  \\
\midrule
GovReport & 17,457 & 972 & 973 \\
SummScreenFD & 3,673 & 338 & 337 \\
QMSum & 1,257 & 272 & 281 \\
Qasper & 2,567 & 1,726 & 1,399 \\
NarrativeQA & 55,003 & 5,878 & 10,306 \\
QuALITY & 2,523 & 2,086 & 2,128 \\
ContractNLI & 7,191 & 1,037 & 2,091 \\
\bottomrule
\end{tabular}
\caption{The number of examples in each train, validation, and test set.}
\label{tab:test_sets_statistics}
\end{table}

\section{Original Datasets Results}

\begin{table}[h]
\small
\centering
\begin{tabular}{@{}llr@{}}
\toprule
\textbf{Dataset} &  \textbf{Metric} &  \textbf{Score} \\
 \midrule
GovReport & ROUGE (1/2) & 56.9 / 22.6  \\
SummScreenFD & ROUGE (1/2) & 25.9 / ~~4.2  \\
QMSum  & ROUGE (1/2) & 29.2 / ~~6.4 \\
Qasper & F1 & 32.8 \\
NarrativeQA & BLEU & 15.53 \\
QuALITY & Acc (Total/Hard) & 30.7 / 29.3 \\
ContractNLI & Acc & 87.5 \\
\bottomrule
\end{tabular}
\caption{Results reported by the datasets' authors, achieved by sequence-to-sequence baselines (when applicable). Most results are \textbf{incomparable} to the results in \scrolls{} as the data was cleaned, filtered and reformatted.}
\label{tab:original_papers_results_col}
\end{table}

\section{Hyperparameters}
\label{appx:hyperparameters}

We finetune each of the baseline models on every dataset separately using AdamW \cite{loshchilov2018decoupled} with $\beta=(\text{0.9}, \text{0.98})$, $\varepsilon=\text{1e-6}$, mixed precision (fp16), and gradient checkpointing.
We achieve an effective batch size of 131,072 ($2^{17}$) tokens by processing 16,384 tokens per GPU across 8 NVIDIA V100 (32GB) GPUs either in parallel or via gradient accumulation.
The summarization datasets are trained for 10 epochs, while Qasper, QuALITY, and ContractNLI are trained for 20; NarrativeQA (the largest dataset) is trained for 2 epochs.
We tune the maximum learning rate over each validation set, selecting from 6 possible values: 1e-5, 2e-5, 5e-5, 1e-4, 2e-4, 5e-4. 
The learning rate is warmed up from zero during the first 10\% of the learning schedule, and then linearly decays back to zero throughout the remaining 90\%. 
We also apply 0.1 dropout throughout each network.
During inference, we generate outputs using greedy decoding.

\begin{table*}[p]
\small
\centering
{\setlength{\extrarowheight}{3pt}%
\begin{tabular}{|p{0.06\linewidth}p{0.885\linewidth}|} 
\hline
\multicolumn{2}{|l|}{\cellcolor[HTML]{ffffff}\textbf{GovReport} \qquad \textit{Summarization}} \\
\cellcolor[HTML]{a0c4e7} \textit{Input}  & 
\cellcolor[HTML]{a0c4e7}
Introduction

The United States has an abundance of natural resources. For much of the nation's history, energy availability was not a concern as commerce and industry needs could be met by domestic supplies. However, industrialization and population growth, and the continuing development of a consumer-oriented society, led to growing dependence...  \vspace{2pt} \\

\cellcolor[HTML]{fcdcdf} \textit{Output}  & 
\cellcolor[HTML]{fcdcdf}
Energy is crucial to the operation of a modern industrial and services economy. Concerns about the availability and cost of energy and about environmental impacts of fossil energy use have led to the establishment of... \vspace{2pt} \\

\hline
\multicolumn{2}{|l|}{\cellcolor[HTML]{ffffff}\textbf{SummScreenFD} \qquad \textit{Summarization}} \\
\cellcolor[HTML]{a0c4e7} \textit{Input}  & 
\cellcolor[HTML]{a0c4e7}
Ted's kitchen

Ted from 2030: Kids, when it comes to love, the best relationships are the ones that just come naturally.

Ted: My first solo batch.

Victoria: Um, I think those need to stay in the oven a while longer. Here's a professional tip. If it's still runny, it's not a cupcake. It's a beverage...  \vspace{2pt} \\

\cellcolor[HTML]{fcdcdf} \textit{Output}  & 
\cellcolor[HTML]{fcdcdf}
Just as things are going well between Ted and Victoria, the latter is offered a surprising but incredible opportunity to be a fellow at a culinary institute in Germany. As the couple discuss the viability of long-distance... \vspace{2pt} \\

\hline
\multicolumn{2}{|l|}{\cellcolor[HTML]{ffffff}\textbf{QMSum} \qquad \textit{Query-Based Summarization}} \\
\cellcolor[HTML]{a0c4e7} \textit{Input}  & 
\cellcolor[HTML]{a0c4e7}
What did the team discuss during the product evaluation about its feature to solve customers' concerns?

~

Project Manager: Yep. Soon as I get this. Okay. This is our last meeting. Um I'll go ahead...  \vspace{2pt}  \\
\cellcolor[HTML]{fcdcdf} \textit{Output}  & 
\cellcolor[HTML]{fcdcdf}
Generally speaking, the team agreed that the product was intuitive and had successfully incorporated main aims that the team had. The team believed the customers were not likely to lose the remote control since it was... \vspace{2pt} \\

\hline
\multicolumn{2}{|l|}{\cellcolor[HTML]{ffffff}\textbf{Qasper} \qquad \textit{Question Answering}} \\
\cellcolor[HTML]{a0c4e7} \textit{Input}  & 
\cellcolor[HTML]{a0c4e7}
Which languages are used in the multi-lingual caption model?

~

Introduction

The bilingual lexicon induction task aims to automatically build word translation dictionaries across different languages, which is beneficial for various natural language processing tasks such as cross-lingual information... \vspace{2pt} \\ 

\cellcolor[HTML]{fcdcdf} \textit{Output}  & 
\cellcolor[HTML]{fcdcdf}
German-English, French-English, and Japanese-English \vspace{2pt} \\

\hline
\multicolumn{2}{|l|}{\cellcolor[HTML]{ffffff}\textbf{NarrativeQA} \qquad \textit{Question Answering}} \\
\cellcolor[HTML]{a0c4e7} \textit{Input}  & 
\cellcolor[HTML]{a0c4e7}
What is the first heist that Dignan and Anthony commit?

~

<b>BOTTLE ROCKET</b>

screenplay by Wes Anderson and Owen Wilson

~

<b>EXT. ALLEY. DAY</b>

ANTHONY and DIGNAN walk down an alley behind a convenience
store. Anthony's nineteen. He's got on a... \vspace{2pt} \\
\cellcolor[HTML]{fcdcdf} \textit{Output}  & 
\cellcolor[HTML]{fcdcdf}
As a practice heist they break into Anthony's family's home. \vspace{2pt} \\

\hline
\multicolumn{2}{|l|}{\cellcolor[HTML]{ffffff}\textbf{QuALITY} \qquad \textit{Multiple-Choice Question Answering}} \\
\cellcolor[HTML]{a0c4e7} \textit{Input}  & 
\cellcolor[HTML]{a0c4e7}
Why did the beings come to Earth?

(A) it was the next planet for them to destroy

(B) they wanted all of Earth's resources

(C) they wanted to take over Earth

(D) they were curious about Earth's creatures

~

"Phone Me in Central Park"

By JAMES McCONNELL

~

There should be an epitaph for... \vspace{2pt} \\
\cellcolor[HTML]{fcdcdf} \textit{Output}  & 
\cellcolor[HTML]{fcdcdf}
it was the next planet for them to destroy \vspace{2pt} \\

\hline
\multicolumn{2}{|l|}{\cellcolor[HTML]{ffffff}\textbf{ContractNLI} \qquad \textit{Natural Language Inference}} \\
\cellcolor[HTML]{a0c4e7} \textit{Input}  & 
\cellcolor[HTML]{a0c4e7}
Agreement shall not grant Receiving Party any right to Confidential Information.

~

NON-DISCLOSURE AND CONFIDENTIALITY AGREEMENT

This NON-DISCLOSURE AND CONFIDENTIALITY AGREEMENT (“Agreement”) is made by and between:

(i) the Office of the United Nations High Commissioner... \vspace{2pt} \\
\cellcolor[HTML]{fcdcdf} \textit{Output}  & 
\cellcolor[HTML]{fcdcdf}
Entailment \vspace{2pt} \\

\hline

\end{tabular}}
\caption{An example from each one of the \scrolls{} datasets, shown in the benchmark's text-to-text format. In this illustration, we truncate the examples' inputs and outputs for brevity.}

\label{tab:data_examples}
\end{table*}

\section{Qualitative Analysis}
\label{appx:qualitative_analysis}
We manually analyze examples from each of the datasets in the benchmark demonstrating cases that require contextualizing and synthesizing information over long ranges of text. Figures \ref{fig:govreport_example}, \ref{fig:qmsum_example}, \ref{fig:qasper_example},  \ref{fig:narrativeqa_example} and \ref{fig:contractnli_example}  showcase gold references,  relevant parts from input documents required to generate those references, and queries when exist, from GovReport, QMSum, Qasper, NarrativeQA and ContractNLI. Together with the SummscreenFD example in Figure~\ref{fig:ssfd_example} and the QuALITY example in Figure~\ref{fig:quality_example} they illustrate cases where important information is spread across multiple sections of the inputs.

\begin{figure}[t]
\small
\centering
{\setlength{\extrarowheight}{3pt}%
\begin{tabular}{|p{0.94\columnwidth}|} 
\hline

\textcolor[HTML]{000000}{The layers of the Internet go far beyond the surface content that many can easily access in their daily searches.}

\textcolor[HTML]{3078BE}{...[486 words]...}

\textcolor[HTML]{000000}{Reportedly, officials are continuously working on expanding techniques to deanonymize activity on the Dark Web and identify malicious actors online.} 
\\
\multicolumn{1}{|c|}{\textcolor[HTML]{AFAFAF}{------ \texttt{{Document}} ------}} \\

\textcolor[HTML]{3078BE}{...[346 words]...}

\textcolor[HTML]{AFAFAF}{Many may consider the Internet and World Wide Web (web) to be synonymous; they are not. Rather, the web is one portion of the Internet, and a medium through which information may be accessed.} \textcolor[HTML]{000000}{In conceptualizing the web, some may view it as consisting solely of the websites accessible through a traditional search engine such as Google. However, this content—known as the ``Surface Web''—is only one portion of the web. The Deep Web refers to ``a class of content on the Internet that, for various technical reasons, is not indexed by search engines,'' and thus would not be accessible through a traditional search engine.}

\textcolor[HTML]{3078BE}{...[3,791 words]...}

\textcolor[HTML]{000000}{the FBI has put resources into developing malware that can compromise servers in an attempt to identify certain users of Tor.} \textcolor[HTML]{AFAFAF}{Since 2002, the FBI has reportedly used a ``computer and internet protocol address verifier'' (CIPAV) to ``identify suspects who are disguising their location using proxy servers or anonymity services, like Tor.'' It has been using this program to target ``hackers, online sexual predators, extortionists, and others.''} \textcolor[HTML]{000000}{Law enforcement has also reportedly been working with companies to develop additional technologies to investigate crimes and identify victims on the Dark Web. In addition to developing technology to infiltrate and deanonymize services such as Tor, law enforcement may rely upon more traditional crime fighting techniques; some have suggested that law enforcement can still rely upon mistakes by criminals} 
\textcolor[HTML]{AFAFAF}{ or flaws in technology to target nefarious actors. For instance, in 2013 the FBI took down the Silk Road, then the ``cyber-underworld's largest black market.'' Reportedly, ``missteps'' by the site's operator led to its demise;}

\textcolor[HTML]{3078BE}{...[979 words]...}
\vspace{2pt}
\\
\hline
\end{tabular}}
\caption{An example from GovReport, a dataset of government reports and their expert-written summaries. This example shows the spread of the relevant information in the document, exemplified by the first and last sentences of the summary.}
\label{fig:govreport_example}
\end{figure}

\begin{figure}
\small
\centering
{\setlength{\extrarowheight}{3pt}%
\begin{tabular}{|p{0.94\columnwidth}|} 
\hline

\textcolor[HTML]{000000}{What did the group discuss about budget balancing?} 
\\
\multicolumn{1}{|c|}{\textcolor[HTML]{AFAFAF}{------ \texttt{{Answer}} ------}} \\

\textcolor[HTML]{AFAFAF}{The use of the LCD screen and the advanced chip cost the team half of the expenditure. Due to the budget limit, the team had to abandon some other designs such as the rubber material and the double-curved structure.}
\textcolor[HTML]{000000}{The USB connection was not feasible for now as well. For the location function, a transmitter, a receiver and speakers could be incorporated on a TV instead}
\vspace{2pt}
\\
\multicolumn{1}{|c|}{\textcolor[HTML]{AFAFAF}{------ \texttt{{Meeting Transcript}} ------}} \\

\textcolor[HTML]{3078BE}{...[1,813 words]...}

\textcolor[HTML]{AFAFAF}{Even then as well , um there was no criteria technically defined for a joystick so I've used what I think's appropriate . With any luck that won't mean that we've incurred more cost than we can actually afford to . }
\textcolor[HTML]{000000}{ It blows a lot of our really good ideas kind of slightly to one side , for example the possibility of having a U\_S\_B\_ connection is definitely not viable now . Um .}

\textcolor[HTML]{3078BE}{...[656 words]...}

\textcolor[HTML]{AFAFAF}{Marketing: We don't even have uh speakers here . The \{disfmarker\} like uh we uh \{disfmarker\}} 
\textcolor[HTML]{000000}{what about speakers and transmitters and stuff like that ? Have we factored that in ?}

\textcolor[HTML]{AFAFAF}{Industrial Designer: Mm .}

\textcolor[HTML]{AFAFAF}{Project Manager: Uh no , we haven't , not \{disfmarker\}}

\textcolor[HTML]{000000}{Marketing: Transmitter , receiver , speakers . Plus the extra device itself that's gonna be on a T\_V\_ .}

\textcolor[HTML]{3078BE}{...[4,651 words]...}
\vspace{2pt}
\\
\hline
\end{tabular}}
\caption{An example from QMSum, a query-based-summarization dataset over meeting transcripts. Information relevant for generating the last two sentences in the answer is spread in different locations in the transcript. }
\label{fig:qmsum_example}
\end{figure}

\begin{figure}
\small
\centering
{\setlength{\extrarowheight}{3pt}%
\begin{tabular}{|p{0.94\columnwidth}|} 
\hline
\textcolor[HTML]{000000}{What approaches without reinforcement learning have been tried?} 
\\
\multicolumn{1}{|c|}{\textcolor[HTML]{AFAFAF}{------ \texttt{{Answer}} ------}} \\

\textcolor[HTML]{000000}{classification, regression, neural methods}
\vspace{2pt}
\\
\multicolumn{1}{|c|}{\textcolor[HTML]{AFAFAF}{------ \texttt{{Article}} ------}} \\

\textcolor[HTML]{3078BE}{...[142 words]...}

\textcolor[HTML]{AFAFAF}{The main contributions of this paper are:}

\textcolor[HTML]{000000}{We compare classification and regression approaches}
\textcolor[HTML]{AFAFAF}{and show that classification produces better results than regression but the quality of the results depends on the approach followed to annotate the data labels.}

\textcolor[HTML]{3078BE}{...[1,006 words]...}

\textcolor[HTML]{AFAFAF}{The bottom section of Table TABREF26 shows the results of }
\textcolor[HTML]{000000}{several variants of the neural architecture. The table includes a neural regressor (NNR) and a neural classifier (NNC).}

\textcolor[HTML]{3078BE}{...[1,398 words]...}
\vspace{2pt}
\\
\hline
\end{tabular}}
\caption{An example from the Qasper dataset, which includes question answering over scientific papers. The evidence for the first part of the reference answer appears in the introduction, while the indication that neural models were also experimented with exists further in the document, in a description of the results table. }
\label{fig:qasper_example}
\end{figure}

\begin{figure}[t]
\small
\centering
{\setlength{\extrarowheight}{3pt}%
\begin{tabular}{|p{0.94\columnwidth}|} 
\hline
\textcolor[HTML]{000000}{Whose initials are on the bottom of the burnt letter to Sir Charles?} 
\\
\multicolumn{1}{|c|}{\textcolor[HTML]{AFAFAF}{------ \texttt{{Answer}} ------}} \\
\textcolor[HTML]{000000}{Laura Lyons}
\vspace{2pt}
\\
\multicolumn{1}{|c|}{\textcolor[HTML]{AFAFAF}{------ \texttt{{Story}} ------}} \\

\textcolor[HTML]{3078BE}{...[35,871 words]...}

\textcolor[HTML]{AFAFAF}{``Well, Sir Henry, your uncle had a letter that morning. He had usually a great many letters, for he was a public man and well known for his kind heart, so that everyone who was in trouble was glad to turn to him. But that morning, as it chanced,}
\textcolor[HTML]{000000}{there was only this one letter, so I took the more notice of it. It was from Coombe Tracey, and it was addressed in a woman's hand.''}

\textcolor[HTML]{AFAFAF}{``Well?''}

\textcolor[HTML]{AFAFAF}{``Well, sir, I thought no more of the matter, and never would have done had it not been for my wife. Only a few weeks ago} \textcolor[HTML]{000000}{she was cleaning out Sir Charles's study--it had never been touched since his death--and she found the ashes of a burned letter in the back of the grate}. \textcolor[HTML]{AFAFAF}{The greater part of it was charred to pieces, but one little slip, the end of a page, hung together, and the writing could still be read, though it was gray on a black ground. It seemed to us to be a postscript at the end of the letter and it said: 'Please, please, as you are a gentleman, burn this letter, and be at the gate by ten o clock.} \textcolor[HTML]{000000}{Beneath it were signed the initials L. L.''}

\textcolor[HTML]{3078BE}{...[861 words]...}

\textcolor[HTML]{AFAFAF}{but among the farmers or gentry there is no one whose initials are those.} \textcolor[HTML]{000000}{Wait a bit though,'' he added after a pause. ``There is Laura Lyons--her initials are L. L.--but she lives in Coombe Tracey.''}

\textcolor[HTML]{3078BE}{...[1,983 words]...}

\textcolor[HTML]{000000}{``Did you ever write to Sir Charles asking him to meet you?'' I continued.}

\textcolor[HTML]{000000}{Mrs. Lyons flushed with anger again. ``Really, sir, this is a very extraordinary question.''}

\textcolor[HTML]{000000}{``I am sorry, madam, but I must repeat it.''}

\textcolor[HTML]{000000}{``Then I answer, certainly not.''}

\textcolor[HTML]{3078BE}{...[97 words]...}

\textcolor[HTML]{AFAFAF}{``You do Sir Charles an injustice. He did burn the letter. But sometimes a letter may be legible even when burned.} \textcolor[HTML]{000000}{You acknowledge now that you
wrote it?''}

\textcolor[HTML]{000000}{``Yes, I did write it,'' she cried, pouring out her soul in a torrent of words. ``I did write it.} \textcolor[HTML]{AFAFAF}{Why should I deny it? I have no reason to be ashamed of it. I wished him to help me.}

\textcolor[HTML]{3078BE}{...[19,996 words]...}
\vspace{2pt}
\\
\hline
\end{tabular}}
\caption{An example from NarrativeQA, where the task is to answer questions about books and movie scripts. In this question about The Hound of the Baskervilles, the answer is first discussed in several places without certainty, where even the final reveal is preceded by an explicit distractor.}
\label{fig:narrativeqa_example}
\end{figure}

\begin{figure}[t]
\small
\centering
{\setlength{\extrarowheight}{3pt}%
\begin{tabular}{|p{0.94\columnwidth}|} 
\hline
\textcolor[HTML]{000000}{All Confidential Information shall be expressly identified by the Disclosing Party.} 
\\
\multicolumn{1}{|c|}{\textcolor[HTML]{AFAFAF}{------ \texttt{{Label}} ------}} \\
\textcolor[HTML]{000000}{Contradiction}
\vspace{2pt}
\\
\multicolumn{1}{|c|}{\textcolor[HTML]{AFAFAF}{------ \texttt{{Contract}} ------}} \\

\textcolor[HTML]{3078BE}{...[427 words]...}

\textcolor[HTML]{000000}{3.1.4 "Confidential Information" means, without limiting the generality of the term: -}

\textcolor[HTML]{AFAFAF}{3.1.4.1 technical, scientific, commercial, financial and market information, trade partners, potential clients, trade leads and trade secrets, and all other information in whatever form, whether in writing or not, whether or not subject to or protected by common law or statutory laws relating to copyright, patent, trademarks, registered or unregistered, or otherwise, disclosed or communicated to the Receiving Party or acquired by the Receiving Party from the Disclosing Party pursuant to this Agreement or the Discussions;}

\textcolor[HTML]{3078BE}{...[1,661 words]...}

 \textcolor[HTML]{000000}{If the Recipient is uncertain as to whether any information is Confidential Information, the Recipient shall treat such information as confidential until the contrary is agreed by the Disclosing Party in writing.}

\textcolor[HTML]{3078BE}{...[4,178 words]...}
\vspace{2pt}
\\
\hline
\end{tabular}}
\caption{An example from ContractNLI, a natural language inference dataset over non-disclosure agreements (NDAs). Here, the challenge of finding the evidence, residing in the middle of a long document, is further amplified by the hypothesis being only implicitly contradicted.}
\label{fig:contractnli_example}
\end{figure}

\end{document}